\documentclass{article}

\usepackage[preprint]{neurips_2025}

\usepackage[utf8]{inputenc} 
\usepackage[T1]{fontenc}    
\usepackage{hyperref}       
\usepackage{url}            
\usepackage{booktabs}       
\usepackage{amsfonts}       
\usepackage{nicefrac}       
\usepackage{microtype}      
\usepackage{xcolor}         

\usepackage{times}
\usepackage{epsfig}
\usepackage{caption}
\usepackage{float}
\usepackage{placeins}
\usepackage{color, colortbl}
\usepackage{stfloats}
\usepackage{enumitem}
\usepackage{tabularx}
\usepackage{xstring}
\usepackage{multirow}
\usepackage{xspace}
\usepackage{url}
\usepackage{subcaption}
\usepackage[hang,flushmargin]{footmisc}
\usepackage{array}
\usepackage{graphicx}
\usepackage{amsmath,amssymb}
\usepackage{marvosym}
\usepackage{makecell}
\usepackage{bm}
\usepackage{wrapfig}
\title{IQE-CLIP: Instance-aware Query Embedding for Zero-/Few-shot Anomaly Detection in Medical Domain}

\newcommand*\samethanks[1][\value{footnote}]{\footnotemark[#1]}
\author{
{\bfseries Hong Huang$^{1,2}$\thanks{Work done at Westlake University.}}\quad
{\bfseries Weixiang Sun$^{1,3}$\samethanks}\quad
{\bfseries Zhijian Wu$^{1}$}\quad
{\bfseries Jingwen Niu$^{4}$}\quad
{\bfseries Donghuan Lu$^{5}$}\\
{\bfseries Xian Wu$^{5}$}\quad
{\bfseries Yefeng Zheng$^{1}$\thanks{Co-corresponding authors.}}\quad \\
\vspace{1pt}\\
{ $^{1}$Westlake University}\quad
{ $^{2}$Simon Fraser University}\quad
{ $^{3}$University of Notre Dame}\\
{ $^{4}$Shandong University}\quad
{ $^{5}$Tencent Jarvis Lab}\\
}

\begin{document}

\maketitle

\begin{abstract}
Recently, the rapid advancements of vision-language models, such as CLIP, leads to significant progress in zero-/few-shot anomaly detection (ZFSAD) tasks. However, most existing CLIP-based ZFSAD methods commonly assume prior knowledge of categories and rely on carefully crafted prompts tailored to specific scenarios. While such meticulously designed text prompts effectively capture semantic information in the textual space, they fall short of distinguishing normal and anomalous instances within the joint embedding space. Moreover, these ZFSAD methods are predominantly explored in industrial scenarios, with few efforts conducted to medical tasks. 
To this end, we propose an innovative framework for ZFSAD tasks in medical domain, denoted as \textbf{IQE-CLIP}. We reveal that query embeddings, which incorporate both textual and instance-aware visual information, are better indicators for abnormalities. Specifically, we first introduce class-based prompting tokens and learnable prompting tokens for better adaptation of CLIP to the medical domain. Then, we design an instance-aware query module (IQM) to extract region-level contextual information from both text prompts and visual features, enabling the generation of query embeddings that are more sensitive to anomalies. Extensive experiments conducted on six medical datasets demonstrate that IQE-CLIP achieves state-of-the-art performance on both zero-shot and few-shot tasks. We release our code and data at \href{https://github.com/hongh0/IQE-CLIP/}{\emph{https://github.com/hongh0/IQE-CLIP/}}.
\end{abstract}

\section{Introduction}
\label{sec:intro}

Anomaly detection (AD), which aims to identify deviations from normal patterns~\cite{cao2024survey,chandola2009anomaly}, is crucial for a wide range of applications, such as industrial quality control~\cite{bergmann2021mvtec,liu2024real3d,zhu2024towards,wang2024real} and medical diagnosis~\cite{fernando2021deep, wyatt2022anoddpm,xiang2023squid}. Traditional anomaly detection methods often rely on large amounts of labeled data~\cite{cao2024bias,DRA,tao2022deep}, thereby constraining their application in real-world scenarios where annotated data is scarce or absent. To address this issue, many efforts have been made on zero-/few-shot anomaly detection (ZFSAD)~\cite{deng2023anovl,winclip,kirillov2023segment} to distinguish anomalies using no or a few labeled examples from the target domain. An essential factor for the performance of ZFSAD approaches is their generalizability and adaptability across different domains, especially in medical filed where images from different modalities and anatomical regions present substantial variations.

Recently, CLIP \cite{radford2021learning}, trained on millions of natural image-text pairs, has demonstrated great generalizability and effectiveness on various downstream tasks~\cite{khattak2023maple,sain2023clip}. Many ZFSAD methods have been built upon CLIP. Specifically, after mapping images and the text prompts to a joint embedding space through the pretrained feature extractors, anomaly can be determined by measuring semantic distance through cosine similarities between visual and textual features~\cite{winclip}. However, the training objective of CLIP is primarily focused on aligning foreground object category information of images with textual semantics, rather than aligning anomalies with defect-related text, which greatly restricts their generalization capabilities for ZFSAD.

\begin{wrapfigure}{r}{0.5\textwidth}
    \includegraphics[width=0.5\textwidth]{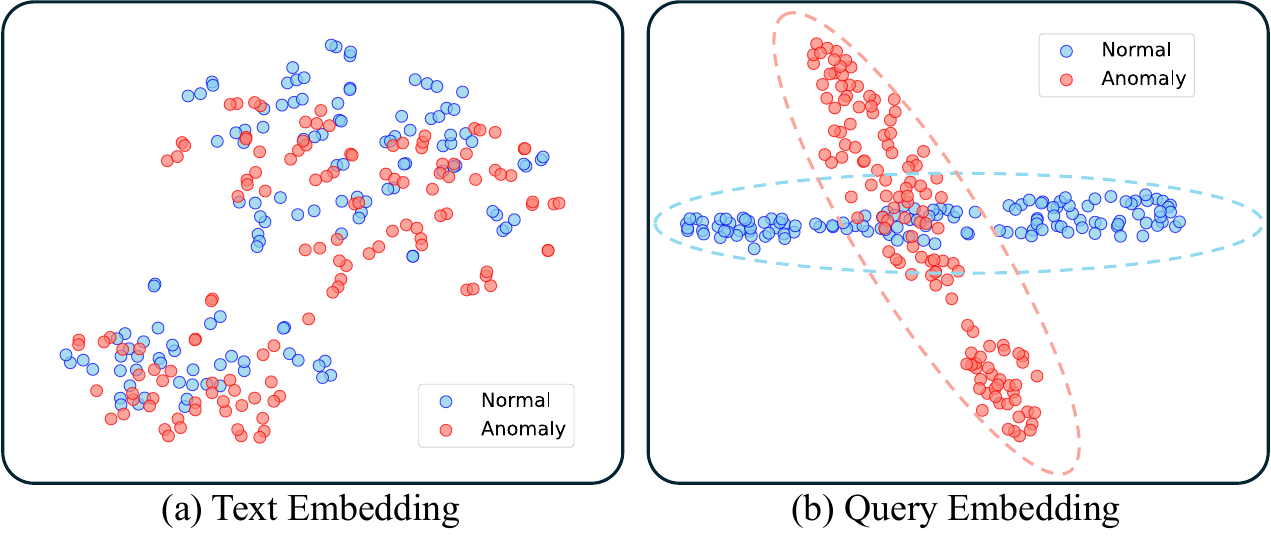}
    \vspace{-1em}
    \captionsetup{font=small}
    \caption{Visualization of text and query embeddings using t-SNE~\cite{JMLR:v9:vandermaaten08a} of the learned text embeddings and query embeddings on the BrainMRI~\cite{menze2014multimodal,baid2021rsna,bakas2017advancing} test set, where \textcolor{red}{red points} represent anomalies, and \textcolor{blue}{blue points} indicate normal instances. 
    The results demonstrate that the features derived from query embeddings effectively separate normal and anomalous instances.}
    \vspace{-1em}
    \label{fig:cmp_tsne}
\end{wrapfigure}
To improve the performance of CLIP-based methods, some previous studies~\cite{winclip,aprilgan,chen2024clip} assume the prior knowledge of categories and elaborately design handcrafted text prompts for better characterization. Other works leverage neural network to learn the text embedding, such as prompt engineering~\cite{cao2023segment, deng2023anovl,li2024promptad} or prompt tuning~\cite{zhou2022learning,sun2022dualcoop,zhou2022conditional}. Furthermore, some apporaches introduce additional linear layers to map visual features into the joint embedding space~\cite{cao2024adaclip,aprilgan,MVFA}, or incorporate post-processing modules to adjust the text embedding based on visual features~\cite{qu2024vcp}. Despite their considerable efforts, several issues remain to impede their applications. First, there is a lack of effective interaction between text and visual features before they are mapped into the joint embedding space, limiting the mutual understanding across different modalities. Second, the adjusted text embedding still fails to distinguish between normal and abnormal instances in the joint embedding space, as shown in Figure~\ref{fig:cmp_tsne} (a). Moreover, meticulously designed text prompts are labor-intensive, and lack of adaptability to diverse scenarios. Last but not the least, these ZFSAD methods are primarily explored in industrial settings, with little research extending them to the medical domain. 

To address these challenges, we propose an instance-aware query embedding model based on CLIP, denoted as IQE-CLIP, focusing on the anomaly detection in medical domain. Firstly, we adopt an innovative prompt tuning strategy by introducing three kinds of tokens: general textual prompting token, class-based prompting token, and learnable prompting token. Through explicitly incorporating category-related visual information and introducing learnable tokens for text encoder, text prompt with better adaptability is constructed for the application on medical domain. In addition, we introduce an instance-aware query module (IQM) to learn query embedding which is more discriminative to anomaly, as shown in Figure~\ref{fig:cmp_tsne} (b). Specifically, a positional-class query is obtained through the image encoder of CLIP along with a learned multilayer perceptron (MLP). Then, the proposed IQM jointly queries multistage visual features for each sample and text embeddings to generate the instance-aware query embedding. To fully verify the proposed IQE-CLIP framework, extensive zero/few-shot experiments are conducted on a challenging medical anomaly detection benchmark~\cite{bao2024bmad}, including images from six different medical modalities and anatomical regions. The contributions of our study can be summarized as:
\begin{itemize}
    \item We show that the instance-aware query embedding, obtained through cross-attention mechanism with instance-wise visual features and text embedding, is more discriminative for anomaly detection tasks.
    \item We introduce an innovative prompt tuning strategy for constructing text prompt with better adaptability. By incorporating class-based prompting tokens and learnable tokens, text embedding, which is more suitable for medical anomaly detection task, can be learned.
    \item We propose an IQM architecture to fully exploit both visual and textual information, leading to query embedding with better characterization ability for anomalies.
    \item Comprehensive experiments in both zero- and few-shot settings are conducted to demonstrate the superiority of the proposed IQE-CLIP framework as well as the effectiveness of each component.
\end{itemize}

\section{Related Work}
\label{sec:related}
\noindent\textbf{Traditional Anomaly Detection.} Traditional anomaly detection methods can be generally divided into two categories, unsupervised~\cite{cao2022informative,roth2022towards,cflowad} and semi-supervised~\cite{cao2024bias,DRA} approaches. Unsupervised methods rely on modeling the normal data distribution with only normal images and then compare the test samples with the learned distribution to determine anomalies~\cite{gong2019memorizing,li2021cutpaste,wu2021learning}. Knowledge distillation~\cite{jiang_masked_2023,MKD}, reconstruction~\cite{DSR,diad,bionda2022deep,yao_feature_2022}, and memory banks\cite{GCPF,regad} are commonly used in these methods for modeling feature distributions. Semi-supervised methods, on the other hand, learn more refined decision boundaries for normal samples by incorporating a small number of anomalous data. With the auxiliary anomalous information, these methods generally outperform the unsupervised ones.

\noindent\textbf{Zero-shot/Few-shot Anomaly Detection.} The pre-trained vision-language model (VLM)~\cite{radford2021learning,kirillov2023segment}, such as CLIP, is widely used for zero-/few-shot anomaly detection. WinCLIP~\cite{winclip} explores its usage by manually designing text prompts for anomaly identification. APRIL-GAN~\cite{aprilgan} additionally introduces a linear layer for image feature adaptation. SAA~\cite{cao2023segment} uses language-guided Grounding DINO~\cite{liu2025grounding} for anomaly region detection, followed by refinement of the detection results with SAM~\cite{kirillov2023segment}. AnomalyCLIP~\cite{zhou2023anomalyclip} enhances text prompts by introducing prompt tuning for adaptation to various scenarios. AdaCLIP~\cite{cao2024adaclip} further strengthens text prompts by introducing hybrid learnable prompts. VCPCLIP~\cite{qu2024vcp} utilizes visual context prompts to fully activate CLIP’s anomaly semantic perception capabilities. Due to the lack of effective interaction with visual features, there is still much room for improvement in existing methods.

\noindent\textbf{Medical Anomaly Detection.}
In medical domain, anomaly detection with few labeled abnormal samples is more challenging due to the large gap between different modalities and anatomical region~\cite{ding2022unsupervised}.
Most methods employ CLIP as base model and expand it with a large set of normal medical scans~\cite{bao2024bmad,cai2023dual,zhou2020encoding}. For example, MedCLIP~\cite{wang2022medclip} adapts CLIP for medical image classification by training on unpaired medical images and text. MediCLIP~\cite{zhang2024mediclip} reduces the requirement of medical images through self-supervised fine-tuning. MVFA~\cite{MVFA} introduces multi-level adaptation into the visual encoder to further improve anomaly detection performance on medical images. Comparing with natural image domains, the efforts made for medical anomaly detection are insufficient and the performance is unsatisfactory. 

\section{Methodology}
\label{sec:methodology}

\begin{figure*}[t]
    \centering
    \includegraphics[width=\linewidth]{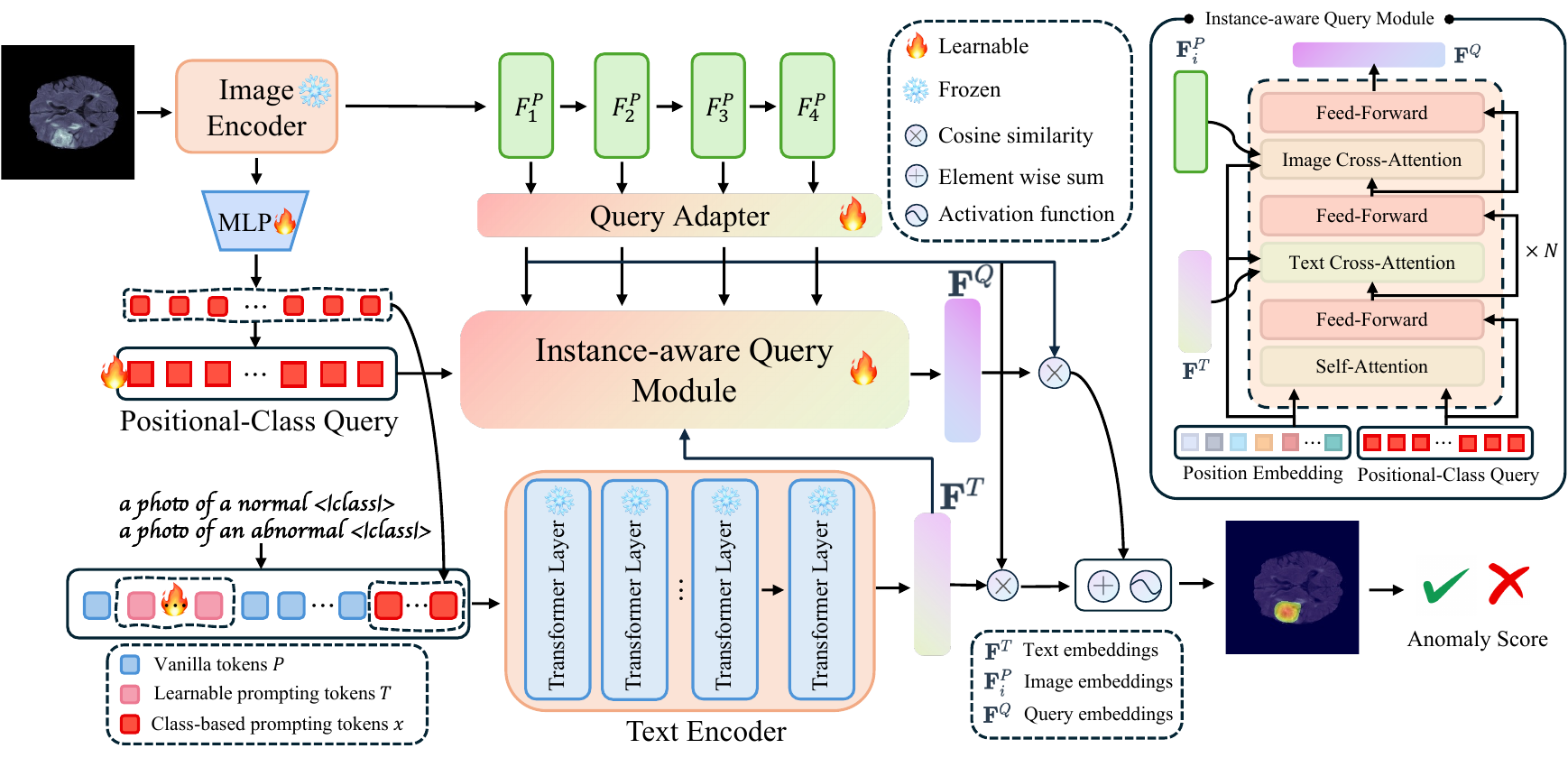}
    \vspace{-1em}
    \caption{Framework of IQE-CLIP, which primarily consists of two steps: prompt tuning and instance-aware query embedding learning. Prompt tuning strategies are introduced to produce more expressive text embeddings, while the IQM leverages cross-modal feature interactions to generate instance-aware query embeddings that are more sensitive to anomalous information.}
    \vspace{-1em}
    \label{fig:framework}
\end{figure*}

\subsection{Problem Definition} \label{subsec:method_problem_def}
Anomaly detection aims to develop a model that associates an input image $I \in \mathbb{R}^{H \times W \times 3}$ with an image-level anomaly score $S$ and a pixel-level anomaly map $\mathbf{M} \in \mathbb{R}^{H \times W}$, indicating whether $I$ and its pixels are normal or anomalous. To address the challenge of adapting a vision-language model originally trained on natural images for anomaly detection in medical domain, we leverage a medical training dataset with annotations for both image-level anomaly classification (AC) and pixel-level anomaly segmentation (AS). Following general zero-/few-shot protocol~\cite{winclip, zhou2023anomalyclip}, we train the network on medical data with pixel-wise annotation from diverse modalities with various anatomical regions. In the zero-shot learning setting, testing data is from unseen modalities and anatomical regions, posing significant challenges for domain generalization. Under the few-shot learning setting, a small subset $I_k^{\text{few}}$, consisting of $K$-labeled images ($K = \{2, 4, 8, 16\}$) within the same modalities and anatomical region as the test set is further applied to enlighten the model with target domain information and facilitate anomaly detection in novel modalities and target regions.

\subsection{Overview} \label{subsec:method_overview}
An overview of our proposed framework is illustrated in Figure \ref{fig:framework}. Given an image $I$, we aim to detect anomalies by computing the similarity in the CLIP embedding space between image and query embeddings as well as the similarity between image and text embeddings. To learn representation with better discriminative ability, the overall framework consists of two main steps, prompt tuning and instance-aware query embedding learning. 
For prompt tuning, we first unify all text prompts by adopting a simple, generic class-agnostic text prompt template. In addition to these static tokens (denoted as vanilla tokens $P$), class-based prompting tokens, which are learned with an MLP by mapping the visual features into query space, are introduced to capture category-related visual information. To further preserve medical domain information for network adaptation, we attach generic learnable prompt tokens to the vanilla tokens in the first $D$ layers of the text encoder. Then, the text encoder, retaining weights of CLIP, is applied to project the prompting tokens to the final text embedding $\mathbf{F}^T$. 
For instance-aware query embedding learning, the IQM is employed to extracts region-level contextual information from text embedding $\mathbf{F}^T$ and layer-wise image embedding $\mathbf{F}_i^P, i \in \{1, 2, 3, 4\}$, obtained by the image encoder. Specifically, the query embedding is first initialized using class query along with specific position embedding. Then, both the text embedding and the image embedding are incorporated through cross-attention mechanisms to gradually obtain the final query embedding $\mathbf{F}^Q$, which presents better characterization for abnormalities. Finally, we integrate the similarities in the CLIP embedding space between image and query embeddings, as well as the ones between image and text embeddings, to compute the anomaly score and anomaly map.

\subsection{Prompt Tuning}
Current CLIP-based ZFSAD methods typically adopt the compositional prompt ensemble (CPE)~\cite{winclip} method to design textual prompts. Specifically, CPE generates textual prompts by combining predefined states and template lists, rather than crafting them freely. Multiple descriptions are created for normal samples, and their features are averaged to obtain the final normal textual representation vector $\mathbf{F}^{T}_N$. The same process is applied to anomalous samples, resulting in the corresponding abnormal vector $\mathbf{F}^{T}_A$. However, the manual design of prompt is labor-intensive and lacks adaptability to diverse scenarios. To address this issue, we propose an innovative prompting strategy with three kinds of tokens: general textual prompting token, class-based prompting token (CPT), and learnable prompting token (LPT).

\noindent\textbf{General Textual Prompting Token.}  
Instead of prompts specifically designed for different products or classes, we first employ a simple class-agnostic general textual prompt, which is defined as:  
\begin{equation}\small
   P = [\text{a}] [\text{photo}] [\text{of}] [\text{a}] [\textit{state}] [\text{class}] [v_1] [v_2] \cdots [v_r],
\end{equation}
where $v_i$ ($i \in \{ 1, 2, \ldots, r \}$) represents a $C$-dimensional learnable vector from the textual embedding space, which is used to learn unified textual context. A pair of contrasting [\textit{state}] words is employed to distinguish textual prompts for normal and abnormal, respectively. For simplicity, we select a common pair of general medical state words, ``\texttt{normal/abnormal},'' tailored to medical datasets.

\noindent\textbf{Class-Based Prompting Tokens (CPT).}  
Due to the lack of consideration for specific contexts, such general textual prompting tokens are inherently insensitive to category-related information, we further introduce CPT to incorporate visual features in a discrete manner.

Specifically, we extract visual features with the pretrained image encoder of CLIP. Leveraging its broad alignment capability of image-text pairs, the visual features embedded at the [CLS] token inherently encode image context and category information. These visual category features, denoted as $x_{\text{cls}}$, are cropped, discretized and then projected to the text embedding space through an MLP.

Finally, the projected visual features are combined with the general textual prompting tokens by adding to the learnable ones, denoted as: 
\begin{equation}\label{eq:cls}
    \{ x_i \}_{i=1}^r = \operatorname{MLP}(x_{\text{cls}}),
\end{equation}
\begin{equation} \label{eq:p}\small
    \quad P_c = [\text{a}][\text{photo}][\text{of}][\text{a}][\textit{state}][\text{class}][x_1 + v_1] \cdots [x_r + v_r],
\end{equation}
where $x_i$ denotes the projected visual feature.
By encoding the category-related information into textual prompting tokens, its adaptability to different image class is effectively improved and the anomaly detection performance is evidently enhanced in diverse scenarios as shown in Sec.~\ref{sec:experiments}.

\noindent\textbf{Learnable Prompting Tokens (LPT).} Due to the lack of medical data during the pretraining of CLIP, the text embedding may lack of discriminability in medical domain. To obtain more suitable text embedding, additional learnable tokens are introduced for the first $D$ layers of text encoder.

Here, we adopt the frozen text encoder of pretrained CLIP for textual embedding learning. The learnable prompting tokens $T \in \mathbb{R}^{M \times C}$ are concatenated to $P_c \in \mathbb{R}^{N \times C}$ as the input for text encoder, where $N$ and $M$ are the lengths of tokens with $M \ll N$. The embedding learning process can be denoted as:  
\begin{equation}
    \left[T_{l+1}, P_{l+1}\right] = \operatorname{TextLayer}_l\left(\left[T_l, P_l\right]\right), \\
\end{equation}

where $\text{TextLayer}_l$ denotes the $l$-th prompting layer, and $C$ represents the embedding dimension.

To improve the refinement of textual space, we sequentially insert learnable token embeddings into the text encoder from lower to upper layers. Specifically, we discard the resulting $T_{l+1}$ of layer $l$ for the first $D$ layers of the text encoder and initialize a new set of learnable token embeddings $T_{l+1}'$ as the input for layer $l+1$ to ensure sufficient calibration. Benefiting from the self-attention mechanism, the output $P_{l+1}$ also contains information learned by $T_l$. To prevent overfitting caused by an excessive number of learnable parameters, we set $D$ with a relatively small value (11 in this study), while the total number of layers in text encoder is 24. For the remaining layers, the results of each layer are retained as the input for the next layer.

\subsection{Instance-aware Query Embedding Learning}
Recent ZFSAD methods typically measure anomaly by computing the cosine similarity between image embeddings $F^P$ with normal text embeddings $\mathbf{F}^{T}_N$ and abnormal ones $\mathbf{F}^{T}_A$. However, without considering the specific information of each image, the text embedding may not be able to provide sufficient guidance to distinguish anomaly, as shown in Figure \ref{fig:cmp_tsne} (a). 
Therefore, we propose to utilize an advanced embedding, denoted as instance-aware query embedding $\mathbf{F}^{Q}\in \mathbb{R}^{2 \times C}$, to better characterize the specific anomaly information. As shown in Figure \ref{fig:cmp_tsne} (b), by jointly querying text embedding as well as visual feature of each image, the query embedding shows much better discriminative ability for anomaly. To obtain query embedding, the proposed IQM incorporates two key designs: class query initialization and multimodal feature interaction.

\noindent\textbf{Positional-Class Query Initialization.} To incorporate positional and class-related information into the query embedding, we first initialize the positional-class query as an input for IQM. Specifically, we leverage the image encoder of CLIP to obtain the global visual feature and take the [CLS] token as the representation of image context and category information. Then, another branch of the MLP is applied to map the image category feature $x_{cls}$ into query space. Furthermore, we concatenate a position embedding with the obtained class query to provide positional guidance, as displayed in the top-right of Figure \ref{fig:framework}.

\noindent\textbf{Multimodal Feature Interaction.}

Due to the significant gap between textual and visual feature spaces, directly measuring the anomaly through cosine similarity between image embedding and text embedding is not accurate enough for anomaly detection. Here, we propose to introduce an instance-aware query module (IQM) for the interaction of text embedding and image embedding. Based on the initialized positional-class query, IQM jointly queries multistage visual features and text embeddings to learn instance-aware query embeddings that are more sensitive to anomalies and better aligned to visual feature space. Specifically, we introduce a query adapter to project the original multistage visual features extracted by CLIP into the query space and then iteratively update the query embedding through cross-attention mechanism, as shown in Figure \ref{fig:framework}.

For an image $I \in \mathbb{R}^{H \times W \times 3}$ , the visual features extracted by CLIP visual encoder are denoted as $F^P_i \in \mathbb{R}^{G \times d}$, where $i \in \{1, 2, 3, 4\}$ , $G$ denotes the size of grids, and $d$ represents the feature dimension. These features are projected into the query space through the query adapter as follows:
\begin{equation}\small
    \mathbf{F}^P_i\in \mathbb{R}^{G \times C} = A_k(\mathrm{ReLU}(F^P_i,W_k)), k \in \{1, \dots, K\},
\end{equation}
where $W_k$ denotes the learnable parameters of a linear transformation, and $A_k$ denotes the query adapter at layer $k$.

In IQM, we first apply self-attention  to the query embedding. Then, cross-attentions are employed with both the image embedding $\mathbf{F}^P_i$ and text embedding $\mathbf{F}^T$. After each attention operation, the embeddings are refined through a feed-forward network (FFN). This query process is repeated several times to produce the final query embedding, which can be expressed as follows:
\begin{equation}\small
\operatorname{Attn}(\mathbf{Q},\mathbf{K},\mathbf{V})=\mathrm{Softmax}(\frac{\mathbf{Q}\cdot\mathbf{K}^\top+\text{PE}}{\sqrt{C}})\cdot\mathbf{V},
\end{equation}
\begin{equation}\small
\mathbf{F}^Q_{l}=\operatorname{FFN}^l(\operatorname{Attn}(\mathbf{F}^Q_{l-1},\mathbf{F}^Q_{l-1},\mathbf{F}^Q_{l-1}), \mathbf{F}^Q_{l-1}),
\end{equation}
\begin{equation}\small
\mathbf{F}^Q_{T}=\operatorname{FFN}^l_T(\operatorname{Attn}^l_T((\mathbf{F}^Q_{l},\mathbf{F}^T,\mathbf{F}^T), \mathbf{F}^Q_{l}),
\end{equation}
\begin{equation}\small
\mathbf{F}^Q_{l+1}=\operatorname{FFN}^l_I(\operatorname{Attn}^l_I(\mathbf{F}^Q_{l},\mathbf{F}^P_i,\mathbf{F}^P_i), \mathbf{F}^Q_{T}),
\end{equation}
where $l$ denotes the number of layer, $\mathbf{Q}$, $\mathbf{K}$,and $\mathbf{V}$ represent the query, key, and value of the attention layer, respectively. Positional embedding (PE)~\cite{NIPS2017_3f5ee243} is introduced for each attention layer. 
The IQM consists of $N$ blocks, with each block comprising three attention layers and three FFN layers.

\noindent\textbf{Anomaly Score Computation.}\
We utilize query embedding $\mathbf{F}^Q$ as well as the text embedding $\mathbf{F}^T$ for the measurement of anomaly. The anomaly map at the $i$-th layer is computed as follows:
\begin{equation}\small
\mathbf{M}_i^\text{query}=\phi\left(\frac{\exp(\cos(\mathbf{\widetilde{F}}_i^P,\mathbf{\widetilde{F}}_A^Q))}{\exp(\cos(\mathbf{\widetilde{F}}_i^P,\mathbf{\widetilde{F}}_N^Q))+\exp(\cos(\mathbf{\widetilde{F}}_i^P,\mathbf{\widetilde{F}}_A^Q))}\right),
\end{equation}
\begin{equation}\small
\mathbf{M}_i^\text{text}=\phi\left(\frac{\exp(\cos(\mathbf{\widetilde{F}}_i^P,\mathbf{\widetilde{F}}_A^T))}{\exp(\cos(\mathbf{\widetilde{F}}_i^P,\mathbf{\widetilde{F}}_N^T))+\exp(\cos(\mathbf{\widetilde{F}}_i^P,\mathbf{\widetilde{F}}_A^T))}\right),
\end{equation}
\begin{equation}\label{eq:sum_map}\small
    \mathbf{M} = \alpha \sum_{i=1}^4 \mathbf{M}_i^\text{query} + (1-\alpha) \sum_{i=1}^4 \mathbf{M}_i^{\text{text}},
\end{equation}
where $\cos(\cdot,\cdot)$ denotes the cosine similarity, $\phi$ represents a deformation and bilinear interpolation upsampling function, and $\widetilde{(\cdot)}$ indicates the L2-normalization along the embedding dimension. The text embeddings $\mathbf{F}^T = [\mathbf{F}^T_N, \mathbf{F}^T_A] \in \mathbb{R}^{2 \times C}$ and query embeddings $\mathbf{F}^Q = [\mathbf{F}^Q_N, \mathbf{F}^Q_A] \in \mathbb{R}^{2 \times C}$ are aligned with the image embedding through the cosine similarity, mapping them to a joint space. This alignment yields the corresponding text anomaly maps $\mathbf{M}_i^\text{text} \in \mathbb{R}^{H \times W}$ and query anomaly maps $\mathbf{M}_i^\text{query} \in \mathbb{R}^{H \times W}$. Subsequently, anomaly maps are extracted from multiple layers in a hierarchical manner~\cite{winclip} and aggregated to produce the final prediction $\mathbf{M}$. A hyperparameter $\alpha$ is introduced to balance the contributions of $\mathbf{M}^\text{text}$ and $\mathbf{M}^\text{query}$.

\subsection{Training and Inference}
\noindent\textbf{Training.} During training, we utilize image-level anomaly annotations $\mathcal{C}$ and pixel-level anomaly annotations $\mathcal{G} \in \mathbb{R}^{H \times W}$ to optimize the image-level anomaly score $S$ and the pixel-level anomaly map $\mathbf{M}$ on the training dataset. Dice loss~\cite{milletari2016v}, focal loss~\cite{ross2017focal}, and binary cross-entropy loss contributes equally to the optimization process as follows:

\begin{equation}\label{eq:general loss}
\mathcal{L}_{i}(\mathbf{M}_i,\mathcal{G},\mathcal{C}) =   \operatorname{Focal}\left(\mathbf{M}_i, \mathcal{G}\right) + \operatorname{Dice}\left(\mathbf{M}_i, \mathcal{G}\right) 
     + \operatorname{BCE}\left(\max\left(\mathbf{M}_i\right), \mathcal{C}\right) ,
\end{equation}

\vspace{-1em}
\begin{equation}\label{eq:sum_loss} \small
    \mathcal{L} = \sum_{i=1}^4 (\alpha \cdot \underbrace{\mathcal{L}_i(\mathbf{M}_i^\text{text}, \mathcal{G}, \mathcal{C})}_{\mathcal{L}_\text{text}} + (1 - \alpha) \cdot \underbrace{\mathcal{L}_i(\mathbf{M}_i^\text{query}, \mathcal{G}, \mathcal{C})}_{\mathcal{L}_\text{query}}).
\end{equation}
The loss function consists of two components: $\mathcal{L}_\text{text}$, derived from the text embeddings, and $\mathcal{L}_\text{query}$, derived from the query embeddings.

\noindent\textbf{Inference.} The final anomaly map $\mathbf{M}$ is obtained by weighted summation of text and query anomaly maps from all stages, as defined in Eq.~\eqref{eq:sum_map}. The anomaly score $S$ is derived by taking the maximum value of anomaly map $\mathbf{M}$.

\section{Experiments}
\label{sec:experiments}

\subsection{Experimental Setups}
\noindent\textbf{Dataset.} We utilize the BMAD~\cite{bao2024bmad} (i.e., Benchmarks for Medical Anomaly Detection) for evaluation, which consists of six datasets spanning five different medical modalities and organs, including Brain MRI, Liver CT, Retinal OCT, Chest X-rays, and Digital His. The BrainMRI~\cite{menze2014multimodal,baid2021rsna,bakas2017advancing}, LiverCT ~\cite{landman2015miccai,bilic2023liver}, and RESC OCT~\cite{hu2019automated} datasets contain both image-level classification labels and pixel-wise anomaly annotations, while OCT17~\cite{kermany2018identifying}, ChestXray~\cite{wang2017chestx}, and HIS~\cite{bejnordi2017diagnostic} only contain image-level classification labels.

\noindent\textbf{Evaluation Metrics. } All performance is evaluated using the area under the receiver operating characteristic (AUC) curve, including image-level AUC for anomaly classification (AC) and pixel-wise AUC for anomaly segmentation (AS).

\subsection{Result Comparison}
\noindent\textbf{Zero-Shot Setting.} The zero-shot AD experiments are performed using a \textit{\textbf{leave-one-out}} setting. In this setup, each target dataset is selected for testing once, while the rest datasets, featuring different modalities and anatomical regions, are used for training, to fully evaluate the generalization ability of the model across different modalities and organs. For BrainMRI, LiverCT and RESC, we measure the performance on anomaly segmentation as well as anomaly classification. For each experiment, we conduct the experiments five times and report the average AUC as well as the standard deviations.

\begin{table*}[t]
\centering
\caption{Comparisons with state-of-the-art \textbf{zero-shot} anomaly detection methods. The AUCs (in $\%$) for AC and AS are reported. The best performance is presented in \textbf{bold}, and the second-best is \underline{underlined}.}
\label{tab:zero}

\rowcolors{2}{white}{gray!5!white} 

{
\setlength{\tabcolsep}{1.0pt}{
\resizebox{\linewidth}{!}{
\begin{tabular}{c|cccccc}
\toprule
Methods & HIS & ChestXray & OCT17 & BrainMRI & LiverCT & RESC \\
\cmidrule(lr){1-7}
WinCLIP~\cite{winclip} & 63.24\footnotesize{$\pm$0.0}~/~- & 43.83\footnotesize{$\pm$0.0}~/~- & 57.04\footnotesize{$\pm$0.0}~/~- & 61.39\footnotesize{$\pm$0.0}~/~39.89\footnotesize{$\pm$0.0} & 42.48\footnotesize{$\pm$0.0}~/~65.04\footnotesize{$\pm$0.0} & 63.67\footnotesize{$\pm$0.0}~/~57.58\footnotesize{$\pm$0.0} \\
APRIL-GAN~\cite{aprilgan} & 67.6\footnotesize{$\pm$2.2}~/~- & 48.59\footnotesize{$\pm$3.5}~/~- & 53.62\footnotesize{$\pm$4.3}~/~- & 63.36\footnotesize{$\pm$3.2}~/~89.53\footnotesize{$\pm$1.6} & 44.19\footnotesize{$\pm$2.6}~/~94.16\footnotesize{$\pm$2.6} & 61.75\footnotesize{$\pm$5.3}~/~81.13\footnotesize{$\pm$4.4} \\
MVFA~\cite{MVFA} & 68.84\footnotesize{$\pm$7.8}~/~- & \underline{66.94\footnotesize{$\pm$7.8}}~/~- & 70.27\footnotesize{$\pm$14.1}~/~- & 68.83\footnotesize{$\pm$9.4}~/~90.30\footnotesize{$\pm$0.4} & \underline{70.44\footnotesize{$\pm$4.9}}~/~\underline{96.86\footnotesize{$\pm$0.5}} & 63.76\footnotesize{$\pm$7.3}~/~87.29\footnotesize{$\pm$0.7} \\
VCPCLIP~\cite{qu2024vcp} & \underline{70.31\footnotesize{$\pm$1.4}}~/~- & \textbf{67.94\footnotesize{$\pm$4.7}}~/~- & \underline{90.91\footnotesize{$\pm$2.0}}~/~- & \underline{73.38\footnotesize{$\pm$4.2}}~/~\underline{90.69\footnotesize{$\pm$1.1}} & 69.95\footnotesize{$\pm$4.5}~/~95.47\footnotesize{$\pm$0.8} & \underline{78.14\footnotesize{$\pm$1.3}}~/~\underline{90.34\footnotesize{$\pm$1.2}} \\
AdaCLIP~\cite{cao2024adaclip} & 65.41\footnotesize{$\pm$4.7}~/~- & 61.46\footnotesize{$\pm$6.6}~/~- & 84.15\footnotesize{$\pm$4.8}~/~- & 67.55\footnotesize{$\pm$3.3}~/~85.76\footnotesize{$\pm$5.3} & 66.69\footnotesize{$\pm$4.8}~/~91.47\footnotesize{$\pm$4.3} & 76.32\footnotesize{$\pm$8.5}~/~71.15\footnotesize{$\pm$15.2} \\
\cmidrule(lr){1-7}
\rowcolor{gray!5!white} 
\textbf{IQE-CLIP (Ours)} & \textbf{74.13\footnotesize{$\pm$3.1}}~/~- & 66.07\footnotesize{$\pm$3.7}~/~- & \textbf{91.39\footnotesize{$\pm$1.1}}~/~- & \textbf{78.95\footnotesize{$\pm$2.3}}~/~\textbf{94.69\footnotesize{$\pm$0.8}} & \textbf{76.86\footnotesize{$\pm$1.7}}~/~\textbf{97.32\footnotesize{$\pm$0.6}} & \textbf{94.15\footnotesize{$\pm$3.4}}~/~\textbf{98.35\footnotesize{$\pm$0.2}} \\
\bottomrule
\end{tabular}}}}
\end{table*}

\begin{table*}[t] 
\centering
\caption{Comparison with state-of-the-art \textbf{few-shot} anomaly detection methods with $K=4$. The AUCs (in $\%$) for anomaly classification (AC) and anomaly segmentation (AS) are reported. The best performance is in \textbf{bold}, and the second-best is \underline{underlined}. \dag~denotes the results reproduced in our experiments, whereas the remaining results are directly extracted from the original papers.}
\label{tab:few}
\small
\setlength{\tabcolsep}{1.7pt}{
\resizebox{\linewidth}{!}{
\begin{tabular}{ccc|ccc|cccccc|c}
\toprule
\multirow{2}{*}{Setting} & \multirow{2}{*}{Method} & \multirow{2}{*}{Source} & HIS & ChestXray & OCT17 & \multicolumn{2}{c}{BrainMRI} & \multicolumn{2}{c}{LiverCT} & \multicolumn{2}{c|}{RESC} & \multirow{2}{*}{Avg.}\\
\cmidrule(lr){4-6}\cmidrule(lr){7-12}
& & & AC & AC & AC & AC & AS & AC & AS & AC & AS &\\
\cmidrule(lr){1-13} 
\multirow{4}{*}{Full-normal-shot} 
& CFlowAD~\cite{cflowad}  & WACV 2022 & 54.54 & 71.44 & 85.43 & 73.97 & 93.52 & 49.93 & 92.78 & 74.43 & 93.75 & 76.64\\
& RD4AD~\cite{RD4AD}  & CVPR 2022 & 66.59 & 67.53 & 97.24 & 89.38 & 96.54 & 60.02 & 95.86 & 87.53 & 96.17 & 84.10\\
& PatchCore~\cite{roth2022towards}  & CVPR 2022 & 69.34 & 75.17 & 98.56 & \underline{91.55} & 96.97 & 60.40 & 96.58 & 91.50 & 96.39 & 86.27\\
& MKD~\cite{MKD} & CVPR 2022 & 77.74 & 81.99 & 96.62 & 81.38 & 89.54 & 60.39 & 96.14 & 88.97 & 86.60 & 84.37\\
\cmidrule(lr){1-13}
\multirow{3}{*}{Few-normal-shot} 
& CLIP~\cite{radford2021learning}  & OpenCLIP & 63.48 & 70.74 & 98.59 & 74.31 & 93.44 & 56.74 & 97.20 & 84.54 & 95.03 & 81.56\\
& MedCLIP~\cite{medclip}  & EMNLP 2022 & 75.89 & \textbf{84.06} & 81.39 & 76.87 & 90.91 & 60.65 & 94.45 & 66.58 & 88.98 & 79.98\\
& WinCLIP~\cite{winclip}  & CVPR 2023 & 67.49 & 70.00 & 97.89 & 66.85 & 94.16 & 67.19 & 96.75 & 88.83 & 96.68 & 82.87\\
\cmidrule(lr){1-13}
\multirow{7}{*}{\makecell[c]{Few-shot}}
& DRA~\cite{DRA}  & CVPR 2022 & 68.73 & 75.81 & 99.06 & 80.62 & 74.77 & 59.64 & 71.79 & 90.90 & 77.28 & 77.62\\
& BGAD~\cite{BGAD}  & CVPR 2023 & - & - & - & 83.56 & 92.68 & 72.48 & \underline{98.88} & 86.22 & 93.84 & 87.94\\
& APRIL-GAN~\cite{aprilgan}  & arXiv 2023 & 76.11 & 77.43 & \textbf{99.41} & 89.18 & 94.67 & 53.05 & 96.24 & 94.70& 97.98 & 86.53\\
& MVFAP$^{\dag}$~\cite{MVFA} & CVPR 2024 & 77.09 & 79.83 & \underline{99.32} & 91.37 & \textbf{97.30} & 78.43 & 98.81 & 94.77 & 98.66 & \underline{90.62}\\
& VCPCLIP$^{\dag}$~\cite{qu2024vcp} & ECCV 2024 & 70.79 & 69.67 & 93.45& 91.18 & 95.02 & \textbf{80.84} & 97.98 & \underline{96.72}& \textbf{98.77} & 88.27\\
& AdaCLIP$^{\dag}$~\cite{cao2024adaclip} & ECCV 2024 & \underline{78.00} & 79.12 & 97.96& 75.44 & 90.56 & 58.14 & 88.17 & 91.15& 97.70 & 84.03\\
\cmidrule(lr){2-13}
&  \multicolumn{2}{c}{\textbf{IQE-CLIP (Ours)}} & \textbf{83.46} & \underline{83.77} & 99.24& \textbf{93.35} & \underline{97.14} & \underline{80.22} & \textbf{99.14} & \textbf{96.87}& \underline{98.76} & \textbf{92.44}\\
\bottomrule
\end{tabular}}
\vspace{-1.5em}
}
\end{table*}

\begin{figure*}[h]
\centering
    \begin{minipage}{0.47\linewidth}
            \includegraphics[width=\linewidth]{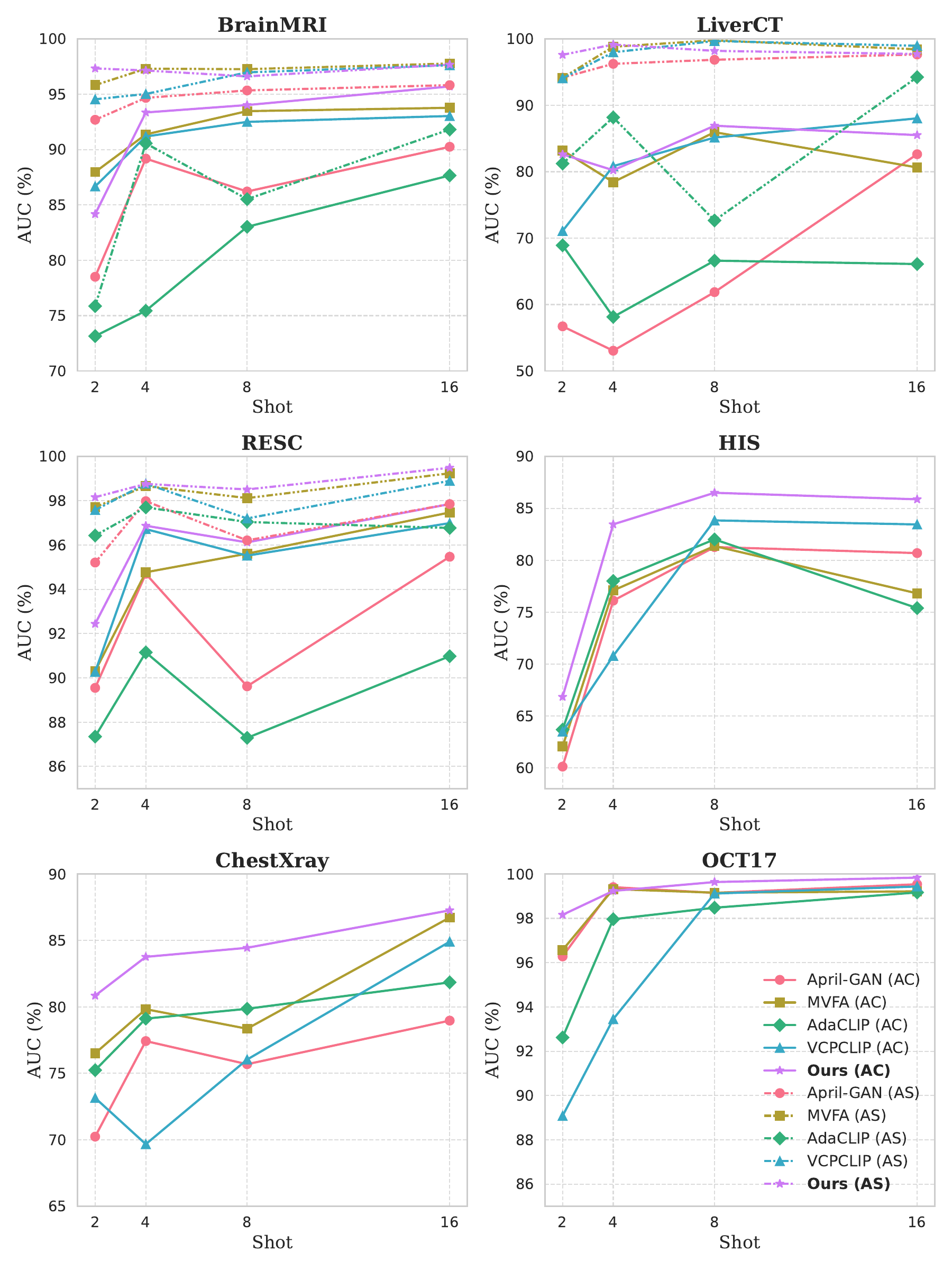}
            \caption{ 
            Comparison with few-shot anomaly detection methods on datasets of BrainMRI, LiverCT,  RESC, HIS,  ChestXray, and OCT17, with the shot number $K = \{2, 4, 8, 16\}$. The average AUCs (in \%) for anomaly classification (AC) and anomaly segmentation (AS) across different shot numbers are reported.
            }
            \vspace{-1em}
            \label{fig:few_fig}
    \end{minipage}
    \hfill
    \begin{minipage}{0.52\linewidth}
        \includegraphics[width = \linewidth]{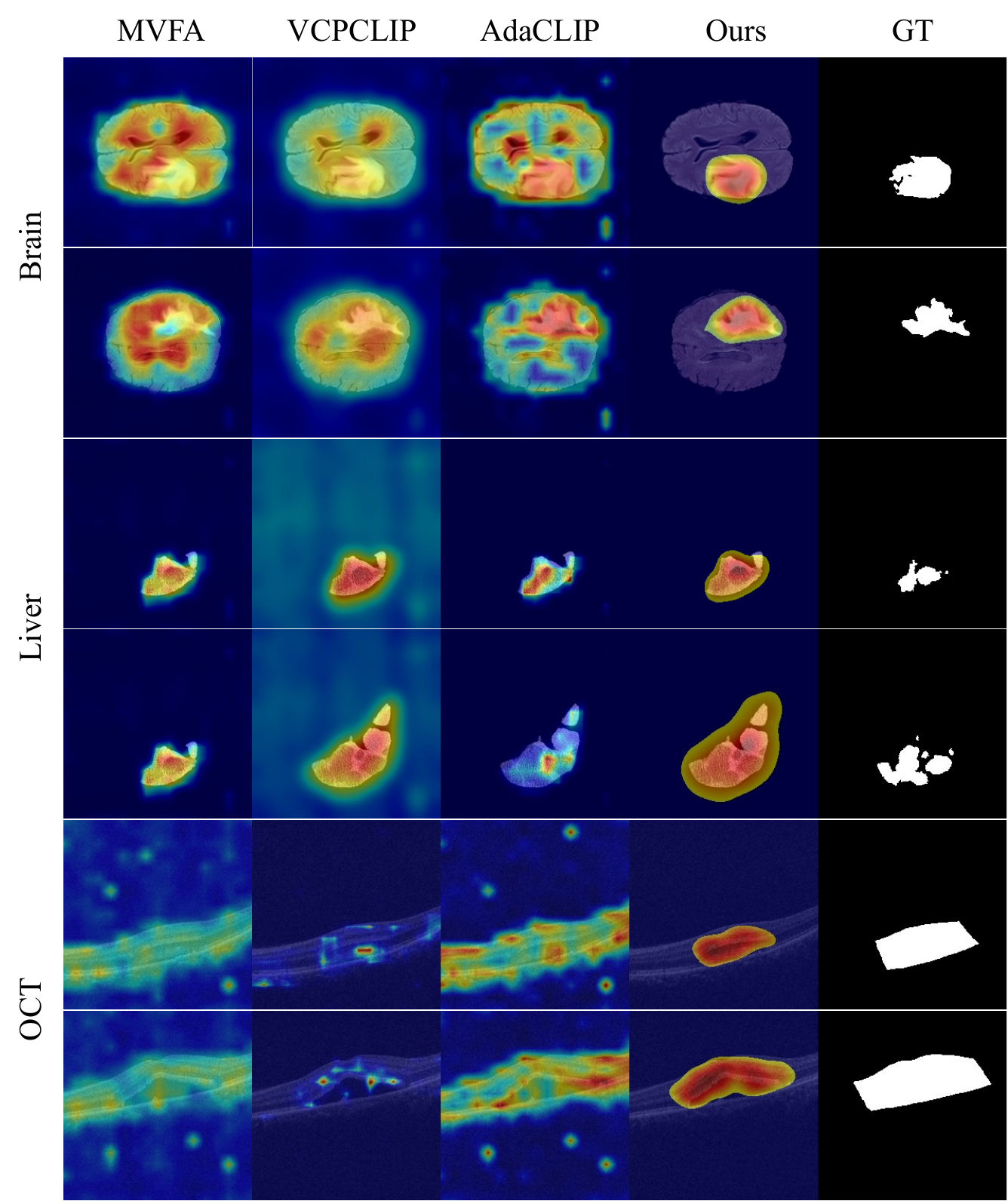}
        \caption{
        Anomaly maps generated by different zero-shot methods on the Brain MRI, Liver CT, and Retinal OCT datasets. From left to right are the visualized results of MVFA, VCPCLIP, AdaCLIP, ours and ground truth, respectively.}
        \vspace{-1em}
        \label{fig:vis}
        
    \end{minipage}

\end{figure*}

As shown in Table~\ref{tab:zero}, our method demonstrates superior performance in different datasets, achieving the best performance in eight out of nine tasks. Compared to recent approaches, our method achieves an average performance improvement of 4.64\% over VCPCLIP, and an improvement of 11\% over AdaCLIP. Furthermore, we observe a substantial reduction in overall variance, decreasing from 2.36\% to 1.88\% compared to the most stable baseline, VCPCLIP, suggesting the effectiveness and reliability of our approach for diverse datasets. In addition, we present a visualization comparison of anomaly maps for different datasets in Figure~\ref{fig:vis}. Our method delivers results that are closer to the true anomalous regions, while the other methods generate incomplete or false-positive results.

\noindent\textbf{Few-Shot Setting.} Table~\ref{tab:few} presents the comparison of our method with SOTA approaches in the few-shot setting with \( K=4 \), while a detailed analysis of performance across various few-shot scenarios can be found in Figure~\ref{fig:few_fig}. Although IQE-CLIP does not achieve the best performance on every task, its average AUC is significantly higher than other approaches (1.82\% higher than the second best method, MVFA). In addition, for the few tasks which IQE-CLIP shows inferior performance, the gap with the best method is less than 0.7\%, suggesting its superiority and robustness across various scenarios.

\subsection{Ablation Study} To thoroughly explore the effectiveness of different components in our model, we conduct ablation studies from three key aspects: (1) components of framework, (2) components of prompting tokens tuning, and (3) components of IQM. The BrainMRI dataset is employed for evaluation and the results are presented in Tables~\ref{tab:abl1}, \ref{tab:abl3}, and \ref{tab:abl2}, respectively.

\begin{wraptable}{r}{0.55\textwidth}
    \vspace{-1em}
    \caption{Ablation study of main components of IQE-CLIP in zero-shot and few-shot settings. The AUCs (in \%) for anomaly classification (AC) and anomaly segmentation (AS) are reported.}
    \label{tab:abl1}
    \small
    \begin{tabular}{c|cc|cc}
    \toprule
        \multirow{2}{*}{Methods} & \multicolumn{2}{c|}{Zero-shot}& \multicolumn{2}{c}{Few-shot} \\
        \cmidrule(lr){2-5}
         &AC  & AS &AC  & AS \\ 
    \midrule
        Baseline & 66.93~& 89.90~& 85.98~& 92.99~ \\ 
        \emph{w/o} prompt tuning&  76.95~& 91.91~& 92.99~  & 95.47~ \\ 
       \emph{w/o} IQM & 75.67~ & 91.99~& 91.79~ & 94.66~  \\ 
        \rowcolor[HTML]{EFEFEF}
        IQE-CLIP & \textbf{78.95}~ & \textbf{94.69}~& \textbf{93.35}~ & \textbf{97.14}~ \\ 
    \bottomrule
    \end{tabular}
    \vspace{-1em}
\end{wraptable}

\noindent\textbf{Components of Framework.}
We first explore the contribution of two main components, prompt tuning and instance-aware query embedding learning. As shown in Table~\ref{tab:abl1}, removing prompt tuning leads to a performance drop of 2\% and 2.78\% in the zero-shot and few-shot settings for AC, respectively, and a drop of 2.78\% and 1.67\% for AS, respectively. Similarly, removing the IQM results in a drop of 3.28\% and 1.56\% in AC for zero-shot and few-shot settings, respectively, and a drop of 2.7\% and 2.48\% in AS, respectively. These findings underscore the critical role of both components. 

\noindent\textbf{Components of Prompt Tuning.} Here, we explore the impact of different prompting tokens, as shown in Table~\ref{tab:abl3}. `Baseline' indicates using only the general textual prompting tokens. The usage of both CPT and LPT leads to significant performance improvement, indicating their criticality for the final text embedding. The combination of both components yields the best performance, suggesting the complementary of CPT and LPT.

\begin{figure*}[]

  \begin{minipage}{0.48\textwidth}
        \centering

 \captionof{table}{Ablation study on the components of the prompting tokens in both zero-shot and few-shot settings. `Baseline' indicates using only the general textual prompting tokens. The AUCs (in \%) for anomaly classification (AC) and anomaly segmentation (AS) are reported.}
        \small
 \label{tab:abl3}
    \setlength\tabcolsep{6pt} 
    \begin{tabular}{c|cc|cc}
\toprule
    \multirow{2}{*}{Methods} & \multicolumn{2}{c|}{Zero-shot}& \multicolumn{2}{c}{Few-shot} \\
    \cmidrule(lr){2-5}
     &AC  & AS &AC  & AS \\ 
\midrule
    Baseline & 66.93~&  89.90~& 85.98~& 92.99~\\
    \emph{w/o} CPT & 74.40~ & 92.80~&90.43~ &95.09~ \\ 
    \emph{w/o} LPT &  74.19~& 92.38~& 91.78~&94.58~ \\ 
    \rowcolor[HTML]{EFEFEF}
    IQE-CLIP & \textbf{78.95}~ &  \textbf{94.69}~&\textbf{93.35}~ &\textbf{97.14}~\\ 
\bottomrule
\end{tabular}%
    \end{minipage} 
    \hfill
    \hfill
\begin{minipage}{0.48\textwidth}
    \centering

    \captionof{table}{Ablation study on the components of IQM in zero-shot and few-shot settings. `Baseline' denotes using the text embedding directly. The AUCs (in \%) for anomaly classification (AC) and anomaly segmentation (AS) are reported.}
    \label{tab:abl2}
    \small
    \setlength\tabcolsep{3pt} 
    \begin{tabular}{c|cc|cc}
    \toprule
        \multirow{2}{*}{Methods} & \multicolumn{2}{c|}{Zero-shot}& \multicolumn{2}{c}{Few-shot} \\
        \cmidrule(lr){2-5}
         &AC  & AS &AC  & AS \\ 
    \midrule
        Baseline & 66.93~& 89.90~& 85.98~& 92.99~\\
        \emph{w/o} class init &  75.34~& 92.72~& 86.07~& 96.17~ \\ 
        \emph{w/o} text embedding&  76.84~& 93.94~&88.09~ & 96.85~ \\ 
        \emph{w/o} image embedding& 76.20~& 92.30~&86.29~ & 95.27~ \\ 
        \rowcolor[HTML]{EFEFEF}
        IQE-CLIP&  \textbf{78.95}~& \textbf{94.69}~& \textbf{93.35}~& \textbf{97.14}~ \\ 
    \bottomrule
    \end{tabular}
	\end{minipage}
        \vspace{-1em}
\end{figure*}

\begin{wrapfigure}{r}{0.45\textwidth}
    \small
    \includegraphics[width=0.45\textwidth]{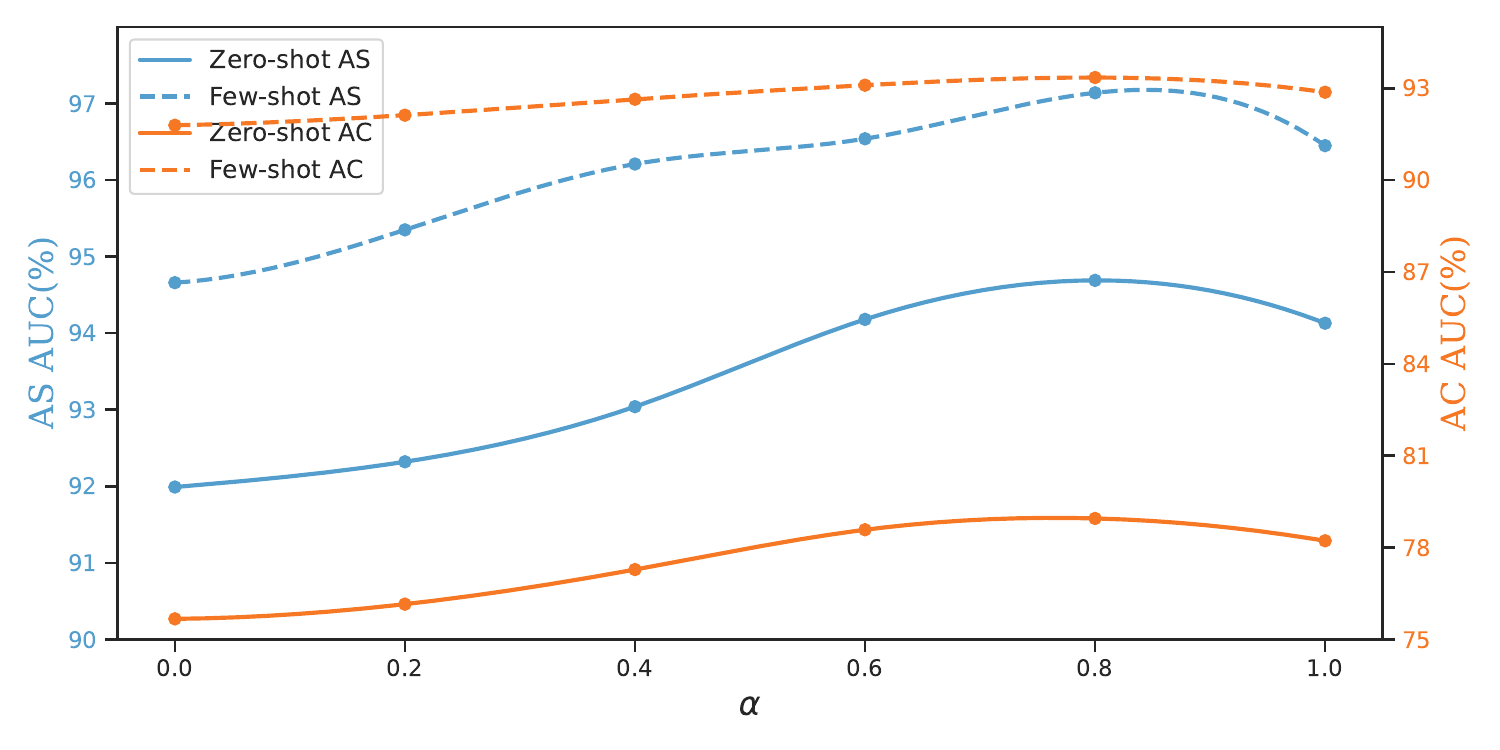}
    \vspace{-2em}
    \captionsetup{font=small}
    \caption{Impact of fusiong weight $\alpha$.}
    \vspace{-1.5em}
    \label{fig:alpha}
\end{wrapfigure}

\noindent\textbf{Components of IQM.} In Table~\ref{tab:abl2}, we present the impact of different components of IQM. `Baseline' denotes using the text embedding directly. The removal of any components yields obvious AUC drop, in which the absence of class initialization leads to the largest degradation, 3.61\% in zero-shot AC and 7.28\% in few-shot AC. When interacting only with visual features or texture features, IQM's performance fails to meet the expectation, indicating that simple interaction with a single modality is insufficient and lack of multimodal information fusion hinders the effectiveness of the query embeddings. On the other hand, interacting with both text embeddings and image embeddings through cross attentions effectively improve the discriminative ability of query embeddings between normal and abnormal instances on both zero and few-shot settings.

\noindent\textbf{Impact of $\alpha$.} In Figure~\ref{fig:alpha}, we present the impact of hyperparemter $\alpha$ in both zero- and few-shot settings, which balances the contribution of query embeddings and text embeddings to anomaly maps. In all experiments, introducing query embeddings leads to better performance than using text embeddings solely, while $\alpha=0.8$ always yields the best results, suggesting that while text embeddings provide valuable prior, query embeddings better characterizes anomaly-specific features for various instances.

\section{Conclusion}
\label{sec:conclusion}
In this paper, we proposed \textbf{IQE-CLIP}, an innovative query embedding-based zero-/few-shot anomaly detection approach for medical scenarios. To learn more discriminative query embeddings for anomaly map computation, two main steps, prompting tuning and query embedding learning, were conducted. By introducing class-based prompting tokens and learnable prompting tokens, text embeddings more adaptable to medical domain were obtained. Then, IQM was introduced to exploit instance-wise visual information as well as the text prompt knowledge through cross attention mechanism. With the learned query embeddings, better distinguishing between normal and abnormal instances can be accomplished. Extensive comparison experiments are performed along with comprehensive ablation studies on six widely used medical datasets in both zero-shot and few-shot settings. The results evidently demonstrated the effectiveness of instance-aware query embeddings for anomaly detection tasks. The performance of our method in other domains will be evaluated in future work.

\bibliographystyle{plain}
\bibliography{references}
\newpage
\appendix

\section{State-of-the-art methods}

\begin{figure*}[h]
\centering
    \begin{minipage}{0.52\linewidth}
            \includegraphics[width=\linewidth]{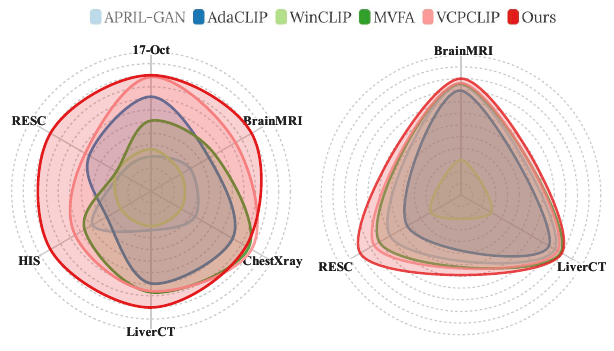}
            \caption{ 
            Quantitative comparison in \textbf{zero-shot} setting with SOTA methods on medical datasets. Left: Radar chart of pixel-level AUC on three datasets with pixel-level annotations. Right: Radar chart of image-level AUC on six datasets with image-level anomaly annotations. 
            }
            \vspace{-1em}
            \label{fig:over_all}
    \end{minipage}
    \hfill
    \begin{minipage}{0.47\linewidth}
        \includegraphics[width = \linewidth]{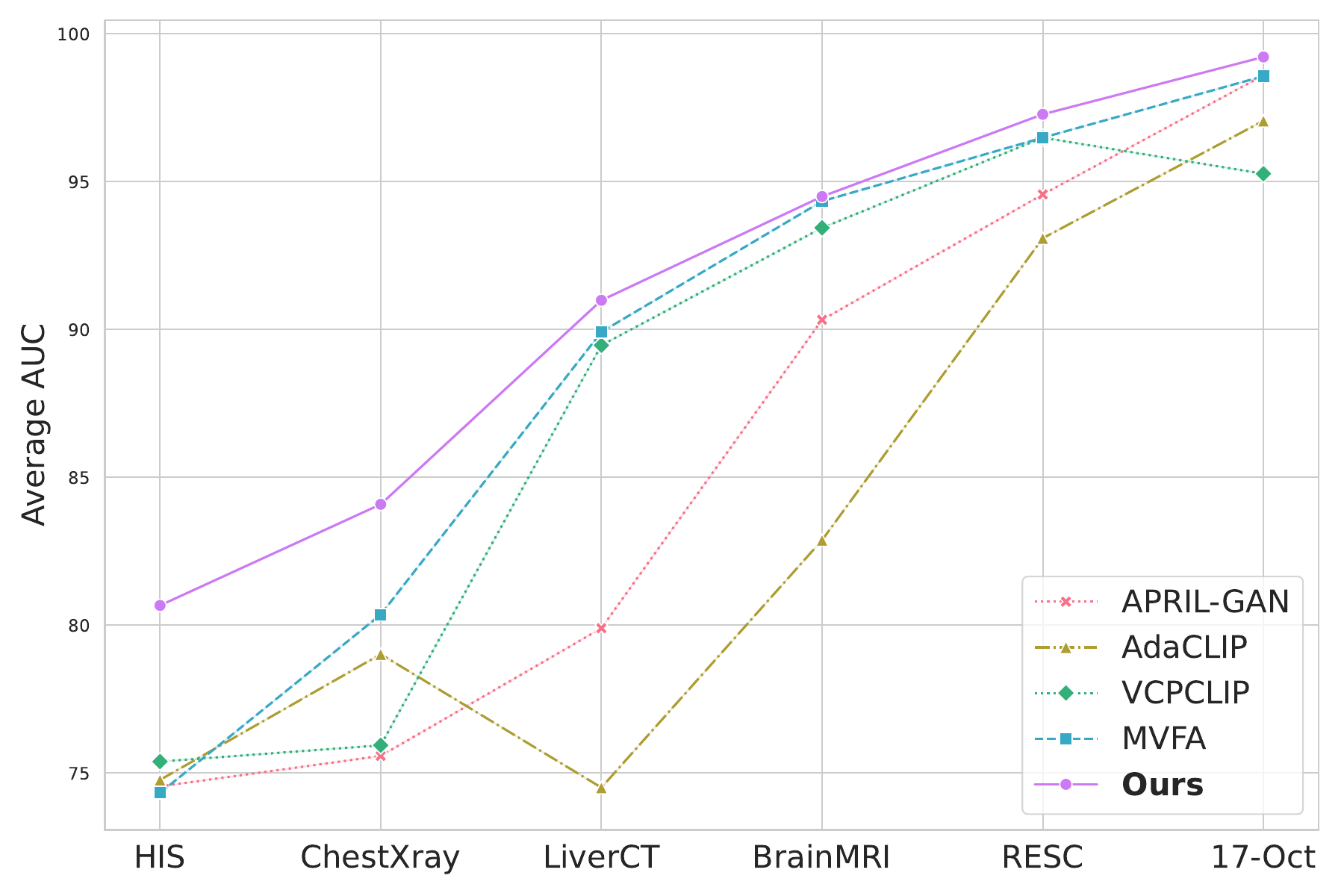}
        \caption{
        Quantitative comparison in \textbf{few-shot} setting with SOTA methods by average AUC on medical datasets, evaluating image-level and pixel-level AUC with shot number $K = \{2, 4, 8, 16\}$. 
        }
        \vspace{-1em}
        \label{fig:few_auc}
        
    \end{minipage}

\end{figure*}

\begin{itemize}
    \item \textbf{WinCLIP}~\cite{winclip} is a method for zero-shot and few-shot anomaly classification and segmentation in industrial quality inspection. WinCLIP improves zero-shot anomaly classification based on CLIP through a compositional prompt ensemble and introduces a window-based CLIP for zero-shot anomaly segmentation that extracts and aggregates multi-scale spatial features aligned with language. WinCLIP by combining language-guided predictions with visual information from normal reference images via a reference association module, enhancing anomaly recognition in few-shot scenarios.

    \item \textbf{APRIL-GAN}~\cite{aprilgan} utilizes the CLIP model to map image features to a joint embedding space for comparison with text features to generate anomaly maps in the zero-shot scenario. Additionally, when reference images are available in the few-shot setting, it employs multiple memory banks to store their features and compares them with the features of test images during testing.

    \item \textbf{MVFA}~\cite{MVFA} proposes a lightweight multi-level adaptation and comparison framework that integrates multiple residual adapters into the pre-trained visual encoder of CLIP to enhance visual features across different levels. It aligns these adapted visual features with text features using multi-level, pixel-wise visual-language feature alignment loss functions, shifting the model's focus from object semantics to anomaly detection in medical images. During testing, it compares the adapted visual features with text prompts and reference image features to generate multi-level anomaly score maps for accurate anomaly classification and segmentation.

    \item \textbf{VCPCLIP}~\cite{qu2024vcp} proposes a visual context prompting model for zero-shot anomaly segmentation, which introduces visual context into text prompts to enhance CLIP's ability to perceive anomalous semantics. It includes a Pre-VCP module that embeds global visual information into text prompts to eliminate the need for product-specific prompts, and a Post-VCP module that adjusts text embeddings based on fine-grained image features to improve segmentation accuracy. 

    \item \textbf{AdaCLIP}~\cite{cao2024adaclip} introduces hybrid learnable prompts to adapt CLIP for detecting anomalies in images from unseen categories without requiring any training samples from these target categories. It combines static prompts shared across all images and dynamic prompts generated for each test image to enhance the model's adaptation capabilities, and further incorporates a Hybrid Semantic Fusion (HSF) module to extract region-level context about anomaly regions, thereby improving image-level anomaly detection performance.

\end{itemize}

\section{Model configuration and training details. }
In this study, we leverage the pre-trained CLIP model (ViT-L-14-336) model from OpenAI~\cite{radford2021learning} as the default backbone. The model comprises 24 layers, which are divided into four stages. The image features used for IQM are extracted from the 6th, 12th, 18th, and 24th layers. All images are resized to $240 \times 240$ pixels for both training and testing.
By default, the learnable text token length $M$ in the prompt tuning layers is set to 4, and the number of layers with learnable promoting tokens $D$ is set to 11. The dimensions of the query embeddings are set to 768. For IQM, the number of attention heads is set to 8, and the number of blocks $N$ is set to 4. The fusion weight $\alpha$ is set to 0.8 by default for predicting the anomaly map. The Adam optimizer is employed with a fixed learning rate of 0.001. Training is conducted for 50 epochs with a batch size of 32.

\section{Datasets details}
\begin{table*}[h]
\small\caption{Datasets details of different medical modalities.}\label{tab:dataset}
\setlength{\tabcolsep}{2.2pt}{
\begin{tabular}{cc|ccccc}
\toprule
     Datasets & Sources & Train (all-normal) & Train (with labels) &  Test& Sample size & Annotation Level \\
    \hline
    BrainMRI & BraTS2021~\cite{baid2021rsna,bakas2017advancing,menze2014multimodal}& 7,500 & 83 & 3,715  &240$\times$240  & Segmentation mask   \\
    LiverCT & BTCV\cite{landman2015miccai} + LiTs~\cite{bilic2023liver} & 1,452 & 166 &1,493  & 512$\times$512 & Segmentation mask  \\
    RESC & RESC~\cite{hu2019automated}& 4,297 & 115 & 1,805  & 512$\times$1,024 & Segmentation mask  \\
    OCT17 & OCT2017~\cite{kermany2018identifying}& 26,315 & 32 &968 & 512$\times$496 & Image label \\
    ChestXray & RSNA~\cite{wang2017chestx}&8,000& 1,490 &17,194 & 1,024$\times$1,024 & Image label    \\
    HIS & Camelyon16~\cite{bejnordi2017diagnostic}&5,088&  236  &2,000 &  256$\times$256 & Image label \\
    \bottomrule 
    \end{tabular}
    }
    
\label{tal:dataset}
\end{table*}

The detailed information of the BMAD~\cite{bao2024bmad} dataset is provided in Table~\ref{tab:dataset}. In the few-shot anomaly detection setting, we randomly select a subset of samples from the labeled training set $I_k^{\text{few}}$, where $K \in \{2, 4, 8, 16\}$. These samples are used for training under the few-shot experimental setup to evaluate the anomaly detection performance of various methods. Below, we provide a detailed description of the BMAD dataset.
 \paragraph{BrainMRI.}This dataset utilizes the latest large-scale brain lesion segmentation dataset, BraTS2021~\cite{baid2021rsna}. To adapt the data for anomaly detection (AD), brain scans and their annotations are axially sliced, selecting only slices that contain a significant portion of brain structures, typically with depths ranging from 60 to 100. Slices without brain tumors are labeled as normal slices. The normal slices from a subset of patients constitute the training set, while the remaining slices are divided into validation and test sets.
\paragraph{LiverCT.}This dataset is primarily constructed from two different datasets: BTCV~\cite{landman2015miccai} and the Liver Tumor Segmentation (LiTS) dataset~\cite{bilic2023liver}. The anomaly-free BTCV dataset is used as the training set. CT scans from the LiTS dataset are utilized to form the evaluation and test sets. For both datasets, the Hounsfield Unit (HU) values of the 3D scans are converted into grayscale using an abdominal window. The scans are then cropped into 2D axial slices, and liver regions are extracted based on the provided segmentation annotations.
\paragraph{Retinal OCT.}This benchmark consists of two OCT datasets: the Retinal Edema Segmentation Challenge (RESC)~\cite{hu2019automated} dataset and the Retinal OCT dataset (OCT2017)~\cite{kermany2018identifying}. The RESC dataset provides pixel-level annotations. The original training, validation, and test sets contain 70, 15, and 15 cases, respectively. Each case consists of 128 slices, some of which exhibit retinal edema. Notably, slices from the same subject appear only in either the validation or test set. OCT2017 is a classification dataset. The normal samples from the original OCT2017 training set are used as training data. In the original test set, images belonging to any of the three disease categories are labeled as anomalies.
\paragraph{ChestXray.}This dataset originates from the development of machine learning (ML) models for chest X-ray diagnosis~\cite{wang2017chestx}. Lung images are associated with nine labels: normal, atelectasis, cardiomegaly, effusion, infiltration, mass, nodule, pneumonia, and pneumothorax, covering eight common thoracic diseases observed in chest X-rays.  
Images belonging to the abnormal categories are labeled as anomalies. The dataset is split into training, test, and validation sets for anomaly detection.
\paragraph{HIS.}This dataset establishes a histopathology benchmark using the Camelyon16 dataset~\cite{bejnordi2017diagnostic}, which focuses on digital pathology imaging for breast cancer metastasis detection. Camelyon16 contains 400 whole-slide images (WSIs) stained with hematoxylin and eosin (H\&E) at 40× magnification, along with multiple lower-resolution versions. Metastasis annotations are provided for the WSIs.  To construct the benchmark dataset, 5,088 normal patches were randomly extracted from 160 normal WSIs in the original Camelyon16 training set for use as training samples. For validation, 100 normal and 100 abnormal patches were extracted from 13 validation WSIs. In the test set, 1,000 normal and 1,000 abnormal patches were extracted from 115 test WSIs in the original Camelyon16 dataset.

\section{Additional experiments}
\begin{table*}[h]
\centering
\caption{Comparisons with SOTA \textbf{few-shot} anomaly detection methods with $K=2,4,8,16$. The average AUCs (in \%) for anomaly classification (AC) are reported. The best result is shown in bold, and the second-best result is underlined.}
\label{tab:suppfew}
\small
\setlength{\tabcolsep}{1.7pt}{
\resizebox{\linewidth}{!}{
\begin{tabular}{ccc|ccc|cccccc}
\toprule
\multirow{2}{*}{Shot Number} & \multirow{2}{*}{Method} & \multirow{2}{*}{Source} & HIS & ChestXray & OCT17 & \multicolumn{2}{c}{BrainMRI} & \multicolumn{2}{c}{LiverCT} & \multicolumn{2}{c}{RESC} \\
\cmidrule(lr){4-6}\cmidrule(lr){7-12}
& & & AC & AC & AC & AC & AS & AC & AS & AC & AS \\
\cmidrule(lr){1-12} 
\multirow{5}{*}{\makecell[c]{2-shot}}
& APRIL-GAN~\cite{aprilgan}  & arXiv 2023 & 60.12 & 70.24 & 96.28 & 78.51 & 92.69 & 56.72 & \underline{94.16} & 89.55 & 95.21 \\
& MVFA~\cite{MVFA} & CVPR 2024 & 62.05 & \underline{76.51} & \underline{96.57} & \textbf{87.97} & \underline{95.82} & \textbf{83.21} & 94.12 & \underline{90.31} & \underline{97.71} \\
& VCPCLIP~\cite{qu2024vcp} & ECCV 2024 & 63.49 & 73.15 & 89.08 & \underline{86.67} & 94.54 & 71.08 & 94.06 & 90.27 & 97.58 \\
& AdaCLIP~\cite{cao2024adaclip} & ECCV 2024 & \underline{63.68} & 75.24 & 92.62 & 73.15 & 75.86 & 68.92 & 81.22 & 87.35 & 96.43 \\
&  \multicolumn{2}{c|}{\textbf{IQE-CLIP (Ours)}} & \textbf{66.84} & \textbf{80.85} & \textbf{98.16} & 84.17 & \textbf{97.33} & \underline{82.61} & \textbf{97.59} & \textbf{92.44} & \textbf{98.16} \\
\midrule
\multirow{5}{*}{\makecell[c]{4-shot}}
& APRIL-GAN~\cite{aprilgan}  & arXiv 2023 & 76.11 & 77.43 & \textbf{99.41} & 89.18 & 94.67 & 53.05 & 96.24 & 94.70 & 97.98 \\
& MVFA~\cite{MVFA} & CVPR 2024 & 77.09 & \underline{79.83} & \underline{99.32} & \underline{91.37} & \textbf{97.30} & 78.43 & \underline{98.81} & 94.77 & 90.62 \\
& VCPCLIP~\cite{qu2024vcp} & ECCV 2024 & 70.79 & 69.67 & 93.45 & 91.18 & 95.02 & \textbf{80.84} & 97.98 & \underline{96.72} & \textbf{98.77} \\
& AdaCLIP~\cite{cao2024adaclip} & ECCV 2024 & \underline{78.00} & 79.12 & 97.96 & 75.44 & 90.56 & 58.14 & 88.17 & 91.15 & 97.70 \\
&  \multicolumn{2}{c|}{\textbf{IQE-CLIP (Ours)}} & \textbf{83.46} & \textbf{83.77} & 99.24 & \textbf{93.35} & \underline{97.14} & \underline{80.22} & \textbf{99.14} & \textbf{96.87} & \underline{98.76} \\
\midrule
\multirow{5}{*}{\makecell[c]{8-shot}}
& APRIL-GAN~\cite{aprilgan}  & arXiv 2023 & 81.27 & 75.69 & 99.15 & 86.21 & 95.34 & 61.87 & 96.85 & 89.62 & 96.21 \\
& MVFA~\cite{MVFA} & CVPR 2024 & 81.37 & 78.36 & \underline{99.16} & \underline{93.47} & \textbf{97.26} & \underline{85.95} & \textbf{99.78} & \underline{95.61} & \underline{98.12} \\
& VCPCLIP~\cite{qu2024vcp} & ECCV 2024 & \underline{83.84} & 76.04 & 99.12 & 92.50 & \underline{96.97} & 85.14 & \underline{99.67} & 95.52 & 97.20 \\
& AdaCLIP~\cite{cao2024adaclip} & ECCV 2024 & 82.01 & \underline{79.86} & 98.48 & 83.02 & 85.52 & 66.61 & 72.66 & 87.29 & 97.04 \\
&  \multicolumn{2}{c|}{\textbf{IQE-CLIP (Ours)}} & \textbf{86.50} & \textbf{84.45} & \textbf{99.64}& \textbf{94.02} & \underline{96.61} & \textbf{86.92} & 98.19 & \textbf{96.12} & \textbf{98.51} \\
\midrule
\multirow{5}{*}{\makecell[c]{16-shot}}
& APRIL-GAN~\cite{aprilgan}  & arXiv 2023 & 80.69 & 78.97 & \underline{99.54} & 90.25 & 95.82 & 82.63 & 97.65 & 95.47 & 97.85 \\
& MVFA~\cite{MVFA} & CVPR 2024 & 76.81 & \underline{86.73} & 99.22 & \underline{93.77} & \textbf{97.77} & 80.64 & \underline{98.42} & \underline{97.47} & \underline{99.24} \\
& VCPCLIP~\cite{qu2024vcp} & ECCV 2024 & \underline{83.45} & 84.91 & 99.44 & 93.03 & 97.62 & \textbf{88.03} & \textbf{98.95} & 96.99 & 98.90 \\
& AdaCLIP~\cite{cao2024adaclip} & ECCV 2024 & 75.39 & 81.85 & 99.17 & 87.66 & 91.81 & 66.10 & 94.23 & 90.98 & 96.77 \\
&  \multicolumn{2}{c|}{\textbf{IQE-CLIP (Ours)}} & \textbf{85.87} & \textbf{87.27} & \textbf{99.84} & \textbf{95.70} & \underline{97.67} & \underline{85.51} & 97.72 & \textbf{97.85} & \textbf{99.49} \\
\bottomrule
\end{tabular}}}
\end{table*}
Quantitative comparison in the zero-shot setting with SOTA methods is illustrated in Figure~\ref{fig:over_all} and additional few-shot setting comparison is presented in Figure~\ref{fig:few_auc}, suggesting the superiority of the proposed IQE-CLIP method.

Table~\ref{tab:suppfew} presents a detailed quantitative analysis of different methods' AD performance under varying shot samples. This analysis is structured in tabular form, explicitly showcasing the specificity of our method across different shot, where $K \in \{2, 4, 8, 16\}$.  

\newpage

\end{document}